\documentclass[10pt,twocolumn,letterpaper]{article}

\usepackage{iccv}
\usepackage{times}
\usepackage{epsfig}
\usepackage{graphicx}
\usepackage{amsmath}
\usepackage{amssymb}
\usepackage{multirow}
\usepackage{bbding}
\usepackage{pifont}
\usepackage{caption}
\usepackage[pagebackref=true,breaklinks=true,letterpaper=true,colorlinks,bookmarks=false]{hyperref}

\iccvfinalcopy 

\def\iccvPaperID{} 

\begin{document}
	
	\title{DSOD: Learning Deeply Supervised Object Detectors from Scratch}
	
\author{Zhiqiang Shen\thanks{indicates equal contribution.  This work was done when Zhiqiang Shen and Zhuang Liu were interns at Intel Labs China. Jianguo Li is the corresponding author.}~$^{1}$, ~ Zhuang Liu\footnotemark[1]~$^{2}$, ~ Jianguo Li$^3$,  ~ Yu-Gang Jiang$^1$,  ~ Yurong Chen$^3$, ~ Xiangyang Xue$^1$ \\
	{$^1$Fudan University,
	$^2$Tsinghua University,
	$^3$Intel Labs China}\\
	\tt\small \{zhiqiangshen13, ygj, xyxue\}@fudan.edu.cn,
	\tt\small liuzhuangthu@gmail.com \\
	\tt\small \{jianguo.li, yurong.chen\}@intel.com
}
\newcommand{\methodname}{dense convolutional  network}
\newcommand{\methodnamecap}{Dense Convolutional Network}
\newcommand{\methodnameshort}{DenseNet}
\newcommand{\methodnameshorts}{DenseNets}
\newcommand{\methodblock}{dense block}
\newcommand{\methodblockcap}{Dense Block}

\newcommand{\regmethodname}{feature drop}
\newcommand{\regmethodnamecap}{Feature Drop}

\newcommand{\stepsizename}{growth rate}

\newcommand{\conv}[1]{$\left[\begin{array}{ll} \text{1}\times \text{1} \text{ conv}\\ \text{3}\times \text{3} \text{ conv} \end{array}\right] \times \text{#1}$}

\newcommand{\cross}[1]{#1 $\times$ #1}

\newcommand{\feati}{x_i}
\newcommand{\clsfeati}{y_i}
\newcommand{\featk}{x_k}
\newcommand{\clsfeatk}{y_k}
\newcommand{\loss}{L}
\newcommand{\featL}{x_L}
\newcommand{\clsfeat}{y}
\newcommand{\anyxs}{\ensuremath{\mathbf{x}}}
\newcommand{\anyys}{\ensuremath{\mathbf{y}}}

\newcommand{\bx}{\ensuremath{\mathbf{x}}}

\newcommand{\sourcexs}{\ensuremath{\mathbf{x^\mathcal{S}}}}
\newcommand{\sourceys}{\ensuremath{\mathbf{y^\mathcal{S}}}}

\newcommand{\targetxs}{\ensuremath{\mathbf{x^\mathcal{T}}}}
\newcommand{\targetys}{\ensuremath{\mathbf{y^\mathcal{T}}}}
\newcommand{\pseudotargetys}{\ensuremath{\mathbf{\hat{y}^\mathcal{T}}}}

	\maketitle
	
	
	\begin{abstract}
		We present Deeply Supervised Object Detector (DSOD), a framework that can learn object detectors from scratch.
		State-of-the-art object objectors rely heavily on the off-the-shelf networks pre-trained on large-scale classification datasets like ImageNet,
		which incurs learning bias due to the difference on both the loss functions and the category distributions between classification and detection tasks.
		Model fine-tuning for the detection task could alleviate this bias to some extent but not fundamentally.
        Besides, transferring pre-trained models from classification to detection between discrepant domains is even more difficult (e.g. RGB to depth images).
        A better solution to tackle these two critical problems is to train object detectors from scratch, which motivates our proposed DSOD.
        Previous efforts in this direction mostly failed due to much more complicated loss functions and limited training data in object detection.
        In DSOD, we contribute a set of design principles for training object detectors from scratch.
		One of the key findings is that deep supervision, enabled by dense layer-wise connections,
    plays a critical role in learning a good detector.
		Combining with several other principles, we develop DSOD following the single-shot detection (SSD) framework.
		Experiments on PASCAL VOC 2007, 2012 and MS COCO datasets demonstrate that DSOD can achieve better results than the state-of-the-art solutions with much more compact models.
		For instance, DSOD outperforms SSD on all three benchmarks with real-time detection speed, while requires only 1/2 parameters to SSD and 1/10 parameters to Faster RCNN. Our code and models are available at: \url{https://github.com/szq0214/DSOD}.

	\end{abstract}
	
\section{Introduction}

Convolutional Neural Networks (CNNs) have produced impressive performance improvements in many computer vision tasks, such as image classification~\cite{krizhevsky2012imagenet,simonyan2014very,szegedy2015going,he2016deep,huang2016densely}, object detection~\cite{girshick2014rich,girshick2015fast,ren2015faster,li2016r,liu2016ssd,redmon2016you}, image segmentation~\cite{long2015fully,hariharan2015hypercolumns,chen2014semantic,yu2015multi}, etc. 
In the past several years, many innovative CNN network structures have been proposed.
Szegedy {\em{et al.}}~\cite{szegedy2015going} propose an ``Inception'' module which concatenates features maps produced by various sized filters. He {\em{et al.}}~\cite{he2016deep} propose residual learning blocks with skip connections, which enable training {\em{very deep}} networks with more than 100 layers.
Huang {\em{et al.}}~\cite{huang2016densely} propose DenseNets with dense layer-wise connections.
Thanks to these excellent network structures, the accuracy of many vision tasks has been greatly improved.
Among them, object detection is one of the fastest moving areas due to its wide applications in surveillance, autonomous driving, etc.

In order to achieve good performance, most of the advanced object detection systems fine-tune networks
pre-trained on ImageNet~\cite{deng2009imagenet}. This fine-tuning process is also viewed as transfer learning~\cite{oquab2014learning}.
Fine-tuning from pre-trained models has at least two advantages.
First, there are many state-of-the-art deep models publicly available.
It is very convenient to reuse them for object detection.
Second, fine-tuning can quickly generate the final model and requires much less instance-level annotated training data than the classification task.
\begin{figure*}[]
	\centering
	\includegraphics[width=0.80\textwidth]{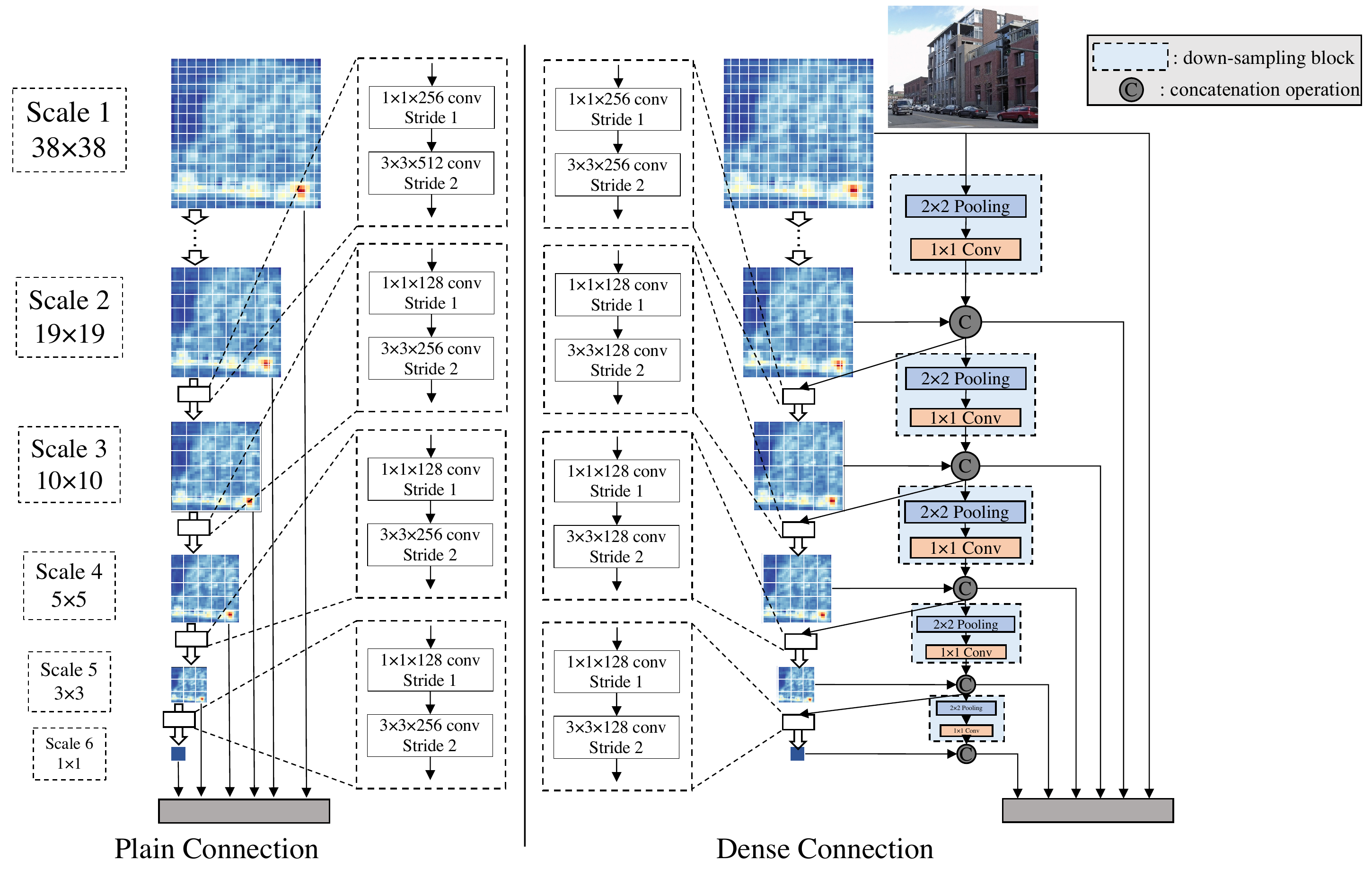}
	\vspace{-0.12in}
	\caption{DSOD prediction layers with plain and dense structures (for 300$\times$300 input). 
		Plain structure is introduced by SSD~\cite{liu2016ssd} and dense structure is ours. 
		See Section~\ref{model} for more details.}
	\label{dense_connection}
	\vspace{-0.16in}
\end{figure*}

However, there are also critical limitations when adopting the pre-trained networks in object detection:
(1) {\em{Limited structure design space}}.
The pre-trained network models are mostly from ImageNet-based classification task, which are usually very heavy --- containing a huge number of parameters.
Existing object detectors directly adopt the pre-trained networks,
and as a result there is little flexibility to control/adjust the network structures (even for small changes of network structure).
The requirement of computing resources is also bounded by the heavy network structures.
(2) {\em{Learning bias}}.
As both the loss functions and the category distributions between classification and detection tasks are different,
we argue that this will lead to different searching/optimization spaces.
Therefore, learning may be biased towards a local minimum which is not the best for detection task.
(3) {\em{Domain mismatch}}. As is known, fine-tuning can mitigate the gap due to different target category distribution.
However, it is still a severe problem when the source domain (ImageNet) has a huge mismatch to the target domain such as depth images, medical images, etc \cite{gupta2016cross}.

Our work is motivated by the following two questions.
First, is it possible to train object detection networks from scratch?
Second, if the first answer is positive,
are there any principles to design a resource efficient network structure for object detection while keeping high detection accuracy?
To meet this goal, we propose deeply supervised objection detectors (DSOD), a simple yet efficient framework which could learn object detectors from scratch.
DSOD is fairly flexible, so that we can tailor various network structures for different computing platforms such as server, desktop, mobile and even embedded devices.

We contribute a set of principles for designing DSOD.
One key point is that {\em{deep supervision}} plays a critical role, which is motivated by the work of~\cite{lee2015deeply,xie2015hed}. In~\cite{xie2015hed}, Xie {\em{et al.}} proposed a holistically-nested structure for edge detection, which included the side-output layers to each conv-stage of base network for explicit deep supervision.  
Instead of using the multiple cut-in loss signals with side-output layers, this paper adopts {\em{deep supervision}} implicitly through the dense layer-wise connections proposed in DenseNet~\cite{huang2016densely}.
Dense structures are not only adopted in the backbone sub-network, but also in the front-end multi-scale prediction layers.
Figure~\ref{dense_connection} illustrates the structure comparison in front-end prediction layers.
The fusion and reuse of multi-resolution prediction-maps help keep or even improve the final accuracy
while reducing model parameters to some extent.

%
Our main contributions are summarized as follows:
\begin{itemize}
	\addtolength{\itemsep}{-0.1in}
	\item[(1)] We present DSOD, to the best of our knowledge, world first framework that can train object detection networks from scratch with state-of-the-art performance.
	\item[(2)] We introduce and validate a set of principles to design efficient object detection networks from scratch through step-by-step ablation studies.
	\item[(3)] We show that our DSOD can achieve state-of-the-art performance on three standard benchmarks (PASCAL VOC 2007, 2012 and MS COCO datasets) with real-time processing speed and more compact models.
\end{itemize}


\renewcommand{\arraystretch}{1.1}
\begin{table*}[]
	\centering
	\resizebox{0.65\textwidth}{!}{%
		\begin{tabular}{c|c|c|c}
			\hline
			\multicolumn{2}{c|}{Layers}                                & Output Size (Input 3$\times$\cross{300}) & DSOD                                                               \\ \hline
			\multirow{4}{*}{Stem}             & Convolution             &64$\times$150$\times$150  & 3$\times$3 conv, stride 2                                                     \\ \cline{2-4}
			& Convolution             & 64$\times$150$\times$150  & 3$\times$3 conv, stride 1                                                     \\ \cline{2-4}
			& Convolution             & 128$\times$150$\times$150 & 3$\times$3 conv, stride 1                                                     \\ \cline{2-4}
			& Pooling                 & 128$\times$75$\times$75   & 2$\times$2 max pool, stride 2                                                \\ \hline
			\multicolumn{2}{c|}{\begin{tabular}[c]{@{}c@{}}Dense Block\\ (1)\end{tabular}}                       & 416$\times$75$\times$75   & \conv{6} \\ \hline
			\multicolumn{2}{c|}{\multirow{2}{*}{\begin{tabular}[c]{@{}c@{}}Transition Layer\\ (1)\end{tabular}}} & 416$\times$75$\times$75   & 1$\times$1 conv                                                              \\ \cline{3-4}
			\multicolumn{2}{c|}{}                                      & 416$\times$38$\times$38   & 2$\times$2 max pool, stride 2                                                \\ \hline
			\multicolumn{2}{c|}{\begin{tabular}[c]{@{}c@{}}Dense Block\\ (2)\end{tabular}}                       & 800$\times$38$\times$38   &                     \conv{8}                                              \\ \hline
			\multicolumn{2}{c|}{\multirow{2}{*}{\begin{tabular}[c]{@{}c@{}}Transition Layer\\ (2)\end{tabular}}} & 800$\times$38$\times$38   & 1$\times$1 conv                                                              \\ \cline{3-4}
			\multicolumn{2}{c|}{}                                      & 800$\times$19$\times$19   & 2$\times$2 max pool, stride 2                                                \\ \hline
			\multicolumn{2}{c|}{\begin{tabular}[c]{@{}c@{}}Dense Block\\ (3)\end{tabular}}                       & 1184$\times$19$\times$19  &                        \conv{8}                                               \\ \hline
			\multicolumn{2}{c|}{Transition w/o Pooling Layer (1)}      & 1184$\times$19$\times$19  & 1$\times$1 conv                                                              \\ \hline
			\multicolumn{2}{c|}{\begin{tabular}[c]{@{}c@{}}Dense Block\\ (4)\end{tabular}}                       & 1568$\times$19$\times$19  &                        \conv{8}                                               \\ \hline
			\multicolumn{2}{c|}{Transition w/o Pooling Layer (2)}      & 1568$\times$19$\times$19  & 1$\times$1 conv                                                              \\ \hline
			\multicolumn{2}{c|}{DSOD Prediction Layers}                           &    --       & Plain/Dense                                                           \\ \hline
		\end{tabular}
	}
	\vspace{-0.05in}
	\caption{DSOD architecture (growth rate $k$ = 48 in each dense block).}
	\label{base_network}
	\vspace{-0.15in}
\end{table*}
\section{Related Work}\label{related_work}
\noindent{\textbf{Object Detection.}}
State-of-the-art CNN based object detection methods can be divided into two groups: (i) region proposal based methods and (ii) proposal-free methods.

Proposal based methods include R-CNN~\cite{girshick2014rich}, Fast R-CNN~\cite{girshick2015fast}, Faster R-CNN~\cite{ren2015faster} and R-FCN~\cite{li2016r}.
R-CNN uses selective search~\cite{uijlings2013selective} to first generate potential object regions in an image and then perform classification on the proposed regions. 
R-CNN requires high computational costs since each region is processed by the CNN network separately. Fast R-CNN and Faster R-CNN improve the efficiency by sharing computation and using neural networks to generate the region proposals. R-FCN further improves speed and accuracy by removing fully-connected layers and adopting position-sensitive score maps for final detection.

Proposal-free methods like YOLO~\cite{redmon2016you} and SSD~\cite{liu2016ssd} have recently been proposed for real-time detection. YOLO uses a single feed-forward convolutional network to directly predict object classes and locations. Comparing with the region-based methods, YOLO no longer requires a second per-region classification operation so that it is extremely fast. SSD improves YOLO in several aspects, including (1) using small convolutional filters to predict categories and anchor offsets for bounding box locations; (2) using pyramid features for prediction at different scales; and (3) using default boxes and aspect ratios for adjusting varying object shapes. Our proposed DSOD is built upon the SSD framework and thus it inherits the speed and accuracy advantages of SSD, while produces smaller and more flexible models.

\noindent{\textbf{Network Architectures for Detection.}}
Significant efforts have been devoted to the design of network architectures for image classification.
Many different networks are emerged, such as AlexNet~\cite{krizhevsky2012imagenet}, VGGNet~\cite{simonyan2014very}, GoogLeNet~\cite{szegedy2015going}, ResNet~\cite{he2016deep} and DenseNet~\cite{huang2016densely}. Meanwhile, several regularization techniques~\cite{srivastava2014dropout,ioffe2015batch} have also been proposed to further enhance the model capabilities.
Most detection methods \cite{girshick2014rich,girshick2015fast,ren2015faster,liu2016ssd} directly utilize pre-trained ImageNet models as the backbone network.

Some other works design specific backbone network structures for object detection, but still require pre-training the network on ImageNet classification dataset first. 
For instance, YOLO~\cite{redmon2016you} defines a network with 24 convolutional layers followed by 2 fully connected layers. YOLO9000~\cite{redmon2016yolo9000} improves YOLO by proposing a new network named Darknet-19, which is a simplified version of VGGNet~\cite{simonyan2014very}.
Kim {\em et al.} \cite{kim2016pvanet} proposes PVANet for object detection, which consists of the simplified ``Inception'' block from GoogleNet.
Huang {\em et al.}~\cite{huang2016speed} investigated various combination of network structures and detection frameworks, and found that Faster R-CNN~\cite{ren2015faster} with Inception-ResNet-v2~\cite{szegedy2016inception} achieved the highest performance.
In this paper, we also consider network structures for generic object detection. 
However, the pre-training on ImageNet is no longer required by the proposed DSOD. 

\noindent{\textbf{Learning Deep Models from Scratch.}}
To the best of our knowledge, there are no works which train object detection networks from scratch.
The proposed approach has very appealing advantages over existing solutions.
We will elaborate and validate the method in the following sections.
%
In semantic segmentation, J{\'e}gou {\em {et al.}}~\cite{jegou2016one} demonstrated that a well-designed network structure can outperform state-of-the-art solutions without using the pre-trained models. It extends DenseNets to fully convolutional networks by adding an upsampling path to recover the full input resolution. 

\section{DSOD}\label{model}
In this section, we first introduce our DSOD architecture and its components, and elaborate several important design principles. Then we describe the training settings.



\subsection{DSOD Architecture}
\paragraph{Overall Framework.}
The proposed DSOD method is a multi-scale proposal-free detection framework similar to SSD \cite{liu2016ssd}.
The network structure of DSOD can be divided into two parts: the backbone sub-network for feature extraction and the front-end sub-network for prediction over multi-scale response maps.
The backbone sub-network is a variant of the deeply supervised DenseNets \cite{huang2016densely} structure, which is composed of a {\em {stem block}}, four {\em {dense blocks}}, two {\em {transition layers}} and two {\em {transition w/o pooling layers}}.
The front-end subnetwork (or named {\em DSOD prediction layers}) fuses multi-scale prediction responses with an elaborated {\em {dense structure}}.
Figure~\ref{dense_connection} illustrates the proposed DSOD prediction layers along with the plain structure of multi-scale predicting maps as used in SSD \cite{liu2016ssd}.
The full DSOD network architecture\footnote{The visualization of the complete network structure is available at: \url{http://ethereon.github.io/netscope/\#/gist/b17d01f3131e2a60f9057b5d3eb9e04d}.} is detailed in Table~\ref{base_network}.
We elaborate each component and the corresponding design principle in the following.

\noindent{\textbf{Principle 1: Proposal-free.}}
We investigated all the state-of-the-art CNN based object detectors, and found that they could be divided into three categories.
First, R-CNN and Fast R-CNN require external object proposal generators like selective search.
Second, Faster R-CNN and R-FCN require integrated region-proposal-network (RPN) to generate relatively fewer region proposals.
Third, YOLO and SSD are single-shot and proposal-free methods, which handle object location and bounding box coordinates as a regression problem.
We observe that only the proposal-free method (the 3rd category) can converge successfully without the pre-trained models.
We conjecture this is due to the RoI (Regions of Interest) pooling in the other two categories of methods ---
RoI pooling generates features for each region proposals, which hinders the gradients being smoothly back-propagated from region-level to convolutional feature maps.
The proposal-based methods work well with pre-trained network models because the parameter initialization is good for those layers before RoI pooling,
while this is not true for training from scratch.

Hence, we arrive at the first principle: training detection network from scratch requires a proposal-free framework.
In practice, we derive a multi-scale proposal-free framework from the SSD framework \cite{liu2016ssd},
as it could reach state-of-the-art accuracy while offering fast processing speed.

\noindent{\textbf{Principle 2: Deep Supervision.}}
The effectiveness of deeply supervised learning has been demonstrated in GoogLeNet~\cite{szegedy2015going}, DSN~\cite{lee2015deeply}, DeepID3~\cite{sun2015deepid3}, etc. 
The central idea is to provide integrated objective function as direct supervision to the earlier hidden layers, rather than only at the output layer. 
These ``companion''  or ``auxiliary'' objective functions at multiple hidden layers can mitigate the ``vanishing'' gradients problem.
The proposal-free detection framework contains both classification loss and localization loss.
The explicit solution requires adding complex side-output layers to introduce ``companion'' objective at each hidden layer for the detection task, similar to \cite{xie2015hed}.
Here we empower \textit{deep supervision} with an elegant \& implicit solution called dense layer-wise connection, as introduced in DenseNets~\cite{huang2016densely}.
A block is called \textit{dense block} when all preceding layers in the block are connected to the current layer. 
Hence, earlier layers in DenseNet can receive additional supervision from the objective function through the skip connections.
Although only a single loss function is required on top of the network, all layers including the earlier layers still can share the supervised signals unencumbered.
We will verify the benefit of deep supervision in Section~\ref{validation}.

\noindent{\emph{Transition w/o Pooling Layer.}}
We introduce this layer in order to increase the number of dense blocks without reducing the final feature map resolution.
In the original design of DenseNet, each transition layer contains a pooling operation to down-sample the feature maps. The number of dense blocks is fixed (4 dense blocks in all DenseNet architectures) if one wants to maintain the same scale of outputs.
The only way to increase network depth is adding layers inside each block for the original DenseNet.
The transition w/o pooling layer eliminates this restriction of the number of dense blocks in our DSOD architecture, and can also be used in the standard DenseNet.

\noindent{\textbf{Principle 3: Stem Block.}}
Motivated by Inception-v3~\cite{szegedy2016rethinking} and v4~\cite{szegedy2016inception}, we define stem block as a stack of three 3$\times$3 convolution layers followed by a 2$\times$2 max pooling layer. The first conv-layer works with stride = 2 and the other two are with stride = 1. We find that adding this simple stem structure can evidently improve the detection performance in our experiments.
We conjecture that, compared with the original design in DenseNet (7$\times$7 conv-layer, stride = 2 followed by a 3$\times$3 max pooling, stride = 2), the stem block can reduce the information loss from raw input images.
We will show that the reward of this stem block is significant for detection performance in Section~\ref{validation}.

\noindent{\textbf{Principle 4: Dense Prediction Structure.}}
Figure~\ref{dense_connection} illustrates the comparison of the plain structure (as in SSD) and our proposed dense structure in the front-end sub-network.
SSD designs prediction-layers as an asymmetric hourglass structure. For 300$\times$300 input images, six scales of feature maps are applied for predicting objects.
The Scale-1 feature maps are from the middle layer of the backbone sub-network, which has the largest resolution (38$\times$38) in order to handle the small objects in an image.
The remaining five scales are on top of the backbone sub-network. Then, a plain transition layer with the {\em {bottleneck}} structure (a 1$\times$1 conv-layer for reducing the number of feature maps plus a 3$\times$3 conv-layer)~\cite{szegedy2016rethinking,he2016deep} is adopted between two contiguous scales of feature maps.

\noindent{\emph{Learning Half and Reusing Half.}}
In the plain structure as in SSD (see Figure~\ref{dense_connection}), each later scale is directly transited from the adjacent previous scale.
We propose the dense structure for prediction, which fuses multi-scale information for each scale.
For simplicity, we restrict that each scale outputs the same number of channels for the prediction feature maps.
In DSOD, in each scale (except scale-1), half of the feature maps are learned from the previous scale with a series of conv-layers,
while the remaining half feature maps are directly down-sampled from the contiguous high-resolution feature maps.
The down-sampling block consists of a 2$\times$2, stride $=$ 2 max pooling layer followed by a 1$\times$1, stride = 1 conv-layer.
The pooling layer aims to match resolution to current size during concatenation.
The 1$\times$1 conv-layer is used to reduce the number of channels to 50\%.
The pooling layer is placed before the 1$\times$1 conv-layer for the consideration of reducing computing cost.
This down-sampling block actually brings each scale with the multi-resolution feature maps from all of its preceding scales,
which is essentially identical to the dense layer-wise connection introduced in DenseNets.
For each scale, we only learn half of new feature maps and reuse the remaining half of the previous ones.
This dense prediction structure can yield more accurate results with fewer parameters than the plain structure, as will be studied in Section~\ref{ablation}.
\begin{table}[]
	\centering
	\resizebox{1\linewidth}{!}{%
		\begin{tabular}{r|cccccccc}
			
			& \multicolumn{7}{c}{DSOD300}                 \\ \hline
			transition w/o pooling? &    & \Checkmark    & \Checkmark  & \Checkmark  & \Checkmark  & \Checkmark  & \Checkmark  & \Checkmark  \\
			hi-comp factor $\theta$?         &  &      & \Checkmark  & \Checkmark  & \Checkmark  & \Checkmark  & \Checkmark  & \Checkmark  \\
			wide bottleneck?   &     &  &      & \Checkmark  & \Checkmark  & \Checkmark  & \Checkmark  & \Checkmark  \\
			wide 1st conv-layer? &    &   &      &      & \Checkmark  & \Checkmark  & \Checkmark  & \Checkmark   \\
			big growth rate?               &    &      &      &      &      & \Checkmark  & \Checkmark  & \Checkmark  \\
			stem block?                           &     &      &      &      &      &      & \Checkmark  & \Checkmark \\
			dense pred-layers?     & &       &      &      &      &      &      & \Checkmark   \\
			\hline
			VOC 2007 mAP                    & 59.9      & 61.6 & 64.5 & 68.6 & 69.7 & 74.5 & 77.3 & 77.7  \\
		\end{tabular}
	}
	\vspace{-0.1in}
	\caption{\textbf{Effectiveness of various designs on VOC 2007 \texttt{test} set.} Please refer to Table~\ref{ablation_study} and Section~\ref{ablation} for more details.}
	\label{effects}
	\vspace{-0.22in}
\end{table}

\renewcommand{\arraystretch}{1.03}
\setlength{\tabcolsep}{1.0em}
\begin{table*}[]
	\centering
	\resizebox{0.84\textwidth}{!}{%
		\begin{tabular}{c|c|c|c|c|c|c|c|c}
			\hline
			Method     & data  & pre-train & transition w/o pooling & stem &     backbone network & prediction Layer & \# parameters & mAP  \\ \hline
			DSOD300   & 07+12 &\ding{55}& \ding{55} &\ding{55}& DS/32-12-16-0.5 &   Plain  & 4.1M & 59.9 \\
			DSOD300   & 07+12 &\ding{55}&\Checkmark&\ding{55}& DS/32-12-16-0.5 &   Plain  & 4.2M & 61.6 \\
			DSOD300   & 07+12 &\ding{55}&\Checkmark&\ding{55}& DS/32-12-16-1 &   Plain  &   5.5M   & 64.5 \\
			DSOD300   & 07+12 &\ding{55}&\Checkmark&\ding{55}&  DS/32-64-16-1  &   Plain & 6.1M & 68.6 \\
			DSOD300   & 07+12 &\ding{55}&\Checkmark&\ding{55}&  DS/64-64-16-1  &   Plain  &   6.3M   & 69.7 \\
			DSOD300   & 07+12 &\ding{55}&\Checkmark&\ding{55}&  DS/64-192-48-1  &   Plain & 18.0M &  74.5\\
			\hline		
			DSOD300   & 07+12 &\ding{55}&\Checkmark&\Checkmark&  DS/64-12-16-1  &   Plain  &   5.2M   & 70.7 \\
			DSOD300   & 07+12 &\ding{55}&\Checkmark&\Checkmark&  DS/64-36-48-1  &   Plain  &   12.5M   & 76.0 \\
			DSOD300   & 07+12 &\ding{55}&\Checkmark&\Checkmark& DS/64-192-48-1  &   Plain  &  18.2M    & 77.3 \\ \hline
			DSOD300   & 07+12 &\ding{55}&\Checkmark&\Checkmark& DS/64-64-16-1  &   Dense &  5.9M  & 73.6 \\
			DSOD300   & 07+12 &\ding{55}&\Checkmark&\Checkmark& DS/64-192-48-1  &   Dense &  14.8M  & 77.7 \\
			DSOD300   & 07+12+COCO &\ding{55}&\Checkmark&\Checkmark& DS/64-192-48-1  &   Dense &  14.8M  & 81.7 \\
			\hline
		\end{tabular}
	}
	\vspace{-0.1in}
	\caption{\textbf{Ablation study on PASCAL VOC 2007 \texttt{test} set.} \textbf{DS/A-B-$k$-$\theta$} describes our backbone network structure. \textbf{A} denotes the number of channels in the 1st conv-layer. \textbf{B} denotes the number of channels in each {\em{bottleneck}} layer (1$\times$1 convolution). \textbf{$k$} is the growth rate in dense blocks. \textbf{$\theta$} denotes the compression factor in transition layers. See Section~\ref{ablation} for more explanations.}
	\label{ablation_study}
	\vspace{-0.1in}
\end{table*}

\subsection{Training Settings}
We implement our detector based on the Caffe framework \cite{jia2014caffe}.
All our models are trained from scratch with SGD solver on NVidia TitanX GPU.
Since each scale of DSOD feature maps is concatenated from multiple resolutions, we adopt the L2 normalization technique~\cite{liu2015parsenet} to scale the feature norm to 20 on all outputs.
Note that SSD only applies this normalization to scale-1. Most of our training strategies follow SSD, including data augmentation, scale and aspect ratios for default boxes and loss function ({\em{e.g.,}} smooth L1 loss for localization purpose and softmax loss for classification purpose), while we have our own learning rate scheduling and mini-batch size settings. Details will be given in the experimental section.

\renewcommand{\arraystretch}{1.03}
\setlength{\tabcolsep}{1.0em}
\begin{table*}[]
	\centering
	\resizebox{1\textwidth}{!}{%
		\begin{tabular}{c|c|c|c|c|c|c|c|c}
			\hline
			Method     & data  & pre-train &    backbone network  & prediction layer & speed (\textit{fps}) & \# parameters & input size & mAP  \\ \hline
			Faster RCNN~\cite{ren2015faster}  & 07+12 &\Checkmark&     VGGNet    &     -      &   7    & 134.7M & $\sim600\times1000$&73.2 \\
			Faster RCNN~\cite{ren2015faster}  & 07+12 &\Checkmark&   ResNet-101    &     -      &  2.4$^*$     &   -   &$\sim600\times1000$& 76.4 \\
			R-FCN~\cite{li2016r}     & 07+12 &\Checkmark&   ResNet-50    &     -      &    11   &31.9M& $\sim600\times1000$& 77.4 \\
			R-FCN~\cite{li2016r}     & 07+12 &\Checkmark&   ResNet-101    &     -      &  9     &50.9M& $\sim600\times1000$& 79.5 \\
			R-FCN{\footnotesize{multi-sc}}~\cite{li2016r}    & 07+12 &\Checkmark&ResNet-101& - & 9   &50.9M&$\sim600\times1000$&  80.5 \\ \hline
			YOLOv2~\cite{redmon2016yolo9000}   & 07+12 &\Checkmark&  Darknet-19       &   -    &     81  &-&$352\times352$&  73.7 \\
			SSD300~\cite{liu2016ssd}   & 07+12 &\Checkmark&  VGGNet       &   Plain    &     46  &26.3M&$300\times300$&  75.8 \\
			SSD300*~\cite{liu2016ssd}   & 07+12 &\Checkmark&   VGGNet       &   Plain    &    46   &26.3M& $300\times300$& 77.2 \\
			\hline
			Faster RCNN  & 07+12 &\ding{55}&  VGGNet/ResNet-101/DenseNet  &   \multicolumn{5}{c}{Failed}    \\
			R-FCN     & 07+12 &\ding{55}&  VGGNet/ResNet-101/DenseNet  &   \multicolumn{5}{c}{Failed}    \\ \hline
			SSD300S$^\dagger$    & 07+12 &\ding{55}&   ResNet-101    &   Plain    &   12.1    &   52.8M   & $300\times300$& 63.8$^*$ \\
			SSD300S$^\dagger$     & 07+12 &\ding{55}&    VGGNet       &   Plain    &   46    &  26.3M    &$300\times300$&  69.6 \\
			SSD300S$^\dagger$     & 07+12 &\ding{55}&    VGGNet       &   Dense    &   37    &  26.0M    &$300\times300$&  70.4 \\ \hline
			DSOD300   & 07+12 &\ding{55}&  DS/64-192-48-1  &   Plain    &  20.6     &  18.2M    &$300\times300$&  77.3 \\
			DSOD300   & 07+12 &\ding{55}&  DS/64-192-48-1  &   Dense   &   17.4    &  14.8M  &$300\times300$&  77.7 \\
			DSOD300   & 07+12+COCO &\ding{55}&  DS/64-192-48-1  &   Dense &   17.4  &  14.8M &$300\times300$& 81.7 \\ \hline
		\end{tabular}
	}
	\vspace{-0.1in}
	\caption{\textbf{PASCAL VOC 2007 \texttt{test} detection results.} SSD300* is updated version by the authors after the paper publication. SSD300S$^\dagger$ indicates training SSD300* from scratch with ResNet-101 or VGGNet, which serves as our baseline. Note that the speed of Faster R-CNN with  ResNet-101 (2.4 \textit{fps}) is tested on K40, while others are tested on Titan X. The result of SSD300S with ResNet-101 (63.8\% mAP, without the pre-trained model) is produced with the default setting of SSD, which may not be optimal.} 
	\label{VOC2007}
	\vspace{-0.2in}
\end{table*}

\renewcommand{\arraystretch}{1.03}
\setlength{\tabcolsep}{1.0em}
\begin{table*}[]\small
	\centering
	\setlength{\tabcolsep}{1.82pt}
	\resizebox{1\textwidth}{!}{%
			\begin{tabular}{l|c|c|c|c|cccccccccccccccccccc}
				\hline
				Method &  data  &  backbone network &  pre-train &  mAP &  aero &  bike &  bird &  boat &  bottle &  bus &  car &  cat &  chair &  cow &  table &  dog &  horse &  mbike &  person &  plant &  sheep &  sofa &  train &  tv \\
				\hline
				ION~\cite{bell2016inside} &  07+12+S &  VGGNet & \Checkmark& 76.4 & 87.5 & 84.7 & 76.8 & 63.8 & 58.3 & 82.6 & 79.0 & 90.9 & 57.8 & 82.0 & 64.7 & 88.9 & 86.5 & 84.7 & 82.3 & 51.4 & 78.2 & 69.2 & 85.2 & 73.5 \\
				Faster RCNN~\cite{ren2015faster} &  07++12 &  ResNet-101 & \Checkmark& 73.8 & 86.5 & 81.6 & 77.2 & 58.0 & 51.0 & 78.6 & 76.6 & 93.2 & 48.6 & 80.4 & 59.0 & 92.1 & 85.3 & 84.8 & 80.7 & 48.1 & 77.3 & 66.5 & 84.7 & 65.6 \\
				R-FCN{\footnotesize{multi-sc}}~\cite{li2016r} &  07++12&  ResNet-101 & \Checkmark& 77.6 & 86.9 & 83.4 & 81.5& 63.8& 62.4 & 81.6 & 81.1 & 93.1 & 58.0 & 83.8 & 60.8& 92.7 & 86.0 & 84.6 & 84.4 & 59.0 & 80.8 & 68.6& 86.1 & 72.9 \\ \hline
				YOLOv2~\cite{redmon2016yolo9000} &  07++12 &  Darknet-19 & \Checkmark& 73.4 & 86.3 & 82.0 & 74.8 & 59.2 & 51.8 & 79.8 & 76.5 & 90.6 & 52.1 & 78.2 & 58.5 & 89.3 & 82.5 & 83.4 & 81.3 & 49.1 & 77.2 & 62.4 & 83.8 & 68.7 \\
				SSD300*~\cite{liu2016ssd} &  07++12&
				VGGNet & \Checkmark& 75.8 & 88.1 & 82.9 & 74.4 & 61.9 & 47.6 & 82.7 & 78.8 & 91.5 & 58.1 & 80.0 & 64.1 & 89.4 & 85.7 & 85.5 & 82.6 & 50.2 & 79.8 & 73.6 & 86.6 & 72.1\\	
				\hline	
				DSOD300 &  07++12&
				DS/64-192-48-1 & \ding{55} & 76.3 & 89.4  & 85.3 & 72.9 & 62.7 & 49.5 & 83.6 & 80.6 & 92.1 & 60.8 & 77.9 & 65.6 & 88.9 & 85.5 & 86.8 & 84.6 & 51.1 & 77.7 & 72.3 & 86.0 & 72.2 \\
				DSOD300 &  07++12+COCO&
				DS/64-192-48-1 & \ding{55} & 79.3 &  90.5 & 87.4 & 77.5 & 67.4 & 57.7 & 84.7 & 83.6 & 92.6 & 64.8 & 81.3 & 66.4 & 90.1 & 87.8 & 88.1 & 87.3 & 57.9 & 80.3 & 75.6 & 88.1 & 76.7 \\
				\hline
			\end{tabular}
	}
	\vspace{-1.6ex}
	\caption{\textbf{PASCAL VOC 2012 \texttt{test} detection results.}
		\textbf{07+12}: 07 \texttt{trainval} + 12 \texttt{trainval}, \textbf{07+12+S}: 07+12 plus segmentation labels, \textbf{07++12}: 07 \texttt{trainval} + 07 \texttt{test} + 12 \texttt{trainval}. Result links are DSOD300 (07+12) : \url{http://host.robots.ox.ac.uk:8080/anonymous/PIOBKI.html}; DSOD300 (07+12+COCO): \url{http://host.robots.ox.ac.uk:8080/anonymous/I0UUHO.html}.} 
	\label{voc12}
	\vspace{-1.5ex}
\end{table*}

\begin{table*}[]
	\centering
	\resizebox{1\textwidth}{!}{%
		\begin{tabular}{l|c|c|c|ccc|ccc|ccc|ccc}
			\hline
			\multirow{2}{*}{Method}          & \multirow{2}{*}{data} & \multirow{2}{*}{network} & \multirow{2}{*}{pre-train} & \multicolumn{3}{c|}{Avg. Precision, IoU:} & \multicolumn{3}{c|}{Avg. Precision, Area:} & \multicolumn{3}{c|}{Avg. Recall, \#Dets:} & \multicolumn{3}{c}{Avg. Recall, Area:} \\
			&                       &              &            & 0.5:0.95        & 0.5        & 0.75       & S            & M            & L            & 1            & 10           & 100         & S           & M           & L           \\ \hline
			Faster RCNN~\cite{ren2015faster} & trainval              & VGGNet          &    \Checkmark     & 21.9            & 42.7       & -          & -            & -            & -            & -            & -            & -           & -           & -           & -           \\
			ION~\cite{bell2016inside}  & train                 & VGGNet    &    \Checkmark   & 23.6            & 43.2       & 23.6       & 6.4          & 24.1         & 38.3         & 23.2         & 32.7         & 33.5        & 10.1        & 37.7        & 53.6        \\
			R-FCN~\cite{li2016r}        & trainval              & ResNet-101 & \Checkmark & 29.2            & 51.5       & -          & 10.3         & 32.4         & 43.3         & -            & -            & -           & -           & -           & -           \\
			R-FCN{\footnotesize{multi-sc}}~\cite{li2016r}        & trainval              & ResNet-101 & \Checkmark & 29.9            & 51.9       & -          & 10.8         & 32.8         & 45.0         & -            & -            & -           & -           & -           & -           \\ \hline
			SSD300 (Huang et al.)~\cite{huang2016speed}        & $<$ trainval35k           & MobileNet   & \Checkmark      & 18.8            & -       & -       & -          & -        & -         & -        & -        & -        & -        & -        & -        \\ 	
			SSD300 (Huang et al.)~\cite{huang2016speed}        & $<$ trainval35k           & Inception-v2   & \Checkmark      & 21.6            & -       & -       & -          & -        & -         & -        & -        & -        & -        & -        & -        \\ 	
			YOLOv2~\cite{redmon2016yolo9000}        & trainval35k           & Darknet-19   & \Checkmark      & 21.6            & 44.0       & 19.2       & 5.0          & 22.4         & 35.5         & 20.7         & 31.6         & 33.3        & 9.8        & 36.5        & 54.4        \\ 		
			SSD300*~\cite{liu2016ssd}        & trainval35k           & VGGNet   & \Checkmark      & 25.1            & 43.1       & 25.8       & 6.6          & 25.9         & 41.4         & 23.7         & 35.1         & 37.2        & 11.2        & 40.4        & 58.4        \\
			DSOD300                           & trainval         & DS/64-192-48-1 & \ding{55} & 29.3            & 47.3       & 30.6       & 9.4          & 31.5         & 47.0         & 27.3         & 40.7         & 43.0        & 16.7        & 47.1        & 65.0        \\ \hline
		\end{tabular}
	}
	\vspace{-1.5ex}
	\caption{\textbf{MS COCO \texttt{test-dev 2015} detection results.}}
	\label{COCO}
	\vspace{-0.18in}
\end{table*}

\section{Experiments}\label{experiments}
We conduct experiments on the widely used PASCAL VOC 2007, 2012 and MS COCO datasets that have 20, 20, 80 object categories respectively.
Object detection performance is measured by mean Average Precision (mAP).

\subsection{Ablation Study on PASCAL VOC2007} \label{ablation}
We first investigate each component and design principle of our DSOD framework.  The results are mainly summarized in Table~\ref{effects} and Table~\ref{ablation_study}.
We design several controlled experiments on PASCAL VOC 2007 with our DSOD300 (with 300$\times$300 inputs) for this ablation study.
A consistent setting is imposed on all the experiments, unless when some components or structures are examined.
In this study, we train the models with the combined training set from VOC 2007 \texttt{trainval} and 2012 \texttt{trainval} (``07+12''), and test on the VOC 2007 testset.

\subsubsection{Configurations in Dense Blocks}
We first investigate the impact of different configurations in dense blocks of the backbone sub-network.

\noindent{\textbf{Compression Factor in Transition Layers.}}
We compare two compression factor values ($\theta$ = 0.5, 1) in the transition layers of DenseNets.
Results are shown in Table~\ref{ablation_study} (rows 2 and 3).
Compression factor $\theta$ = 1 means that there is no feature map reduction in the transition layer, while $\theta$ = 0.5 means half of the feature maps are reduced.
Results show that $\theta$ = 1 yields 2.9\% higher mAP than $\theta$ = 0.5.

\noindent{\textbf{\# Channels in bottleneck layers.}}
As shown in Table~\ref{ablation_study} (rows 3 and 4),
we observe that wider bottleneck layers (with more channels of response maps) improve the performance greatly (4.1\% mAP).

\noindent{\textbf{\# Channels in the 1st conv-layer}}
We observe that a large number of channels in the first conv-layers is beneficial, which brings 1.1\% mAP improvement (in Table~\ref{ablation_study} rows 4 and 5).

\noindent{\textbf{Growth rate.}}
A large growth rate $k$ is found to be much better.
We observe 4.8\% mAP improvement in Table~\ref{ablation_study} (rows 5 and 6) when increase $k$ from 16 to 48 with 4$k$ bottleneck channels.

\vspace{-0.10in}
\subsubsection{Effectiveness of Design Principles} \label{validation}
We now justify the effectiveness of the key design principles elaborated earlier.

\noindent{\textbf{Proposal-free Framework.}}
We tried to train object detectors from scratch using the proposal-based framework such as Faster R-CNN and R-FCN.
However, the training process failed to converge for all the network structures we attempted (VGGNet, ResNet, DenseNet).
We further tried to train object detectors using the proposal-free framework SSD.
The training converged successfully but gives much worse results (69.6\% for VGG), compared with the case fine-tuning from pre-trained model (75.8\%), as shown in Table~\ref{VOC2007}.
This experiment validates our design principle to choose a proposal-free framework.

\noindent{\textbf{Deep Supervision.}}
We then tried to train object detectors from scratch with deep supervision.
Our DSOD300 achieves 77.7\% mAP, which is much better than the SSD300S that
is trained from scratch using VGG16 (69.6\%) without deep supervision.
It is also much better than the fine-tuned results by SSD300 (75.8\%).
This validates the principle of deep supervision.

\noindent{\textbf{Transition w/o Pooling Layer.}}
We compare the case without this designed layer (only 3 dense blocks) and the case with the designed layer (4 dense blocks in our design).
The backbone network is DS/32-12-16-0.5. Results are shown in Table~\ref{ablation_study}. The network structure with the {\em Transition w/o pooling layer} brings 1.7\% performance gain,
which validates the effectiveness of this layer. 

\noindent{\textbf{Stem Block.}}
As can be seen in Table~\ref{ablation_study} (rows 6 and 9), the stem block improves the performance from 74.5\% to 77.3\%. 
This validates our conjecture that using stem block can protect information loss from the raw input images.

\noindent{\textbf{Dense Prediction Structure.}}
We analyze the dense prediction structure from three aspects: speed, accuracy and parameters.
As shown in Table~\ref{VOC2007}, DSOD with dense front-end structure runs slightly lower than the plain structure (17.4 \textit{fps} vs. 20.6 \textit{fps}) on a Titan X GPU, due to the overhead from additional down-sampling blocks.
However, the dense structure improves mAP from 77.3\% to 77.7\%,
while reduces the parameters from 18.2M to 14.8M.
Table~\ref{ablation_study} gives more details (rows 9 and 10).
We also tried to replace the prediction layers in SSD with the proposed dense prediction layers.
The accuracy on VOC 2007 test set can be improved from 75.8\% (original SSD) to 76.1\% (with pre-trained models), and 69.6\% to 70.4\% (w/o pre-trained models), when using the VGG-16 models as backbone.
This verifies the effectiveness of the dense prediction layer.

\noindent{\textbf{What if pre-training on ImageNet?}}
It is interesting to see the performance of DSOD with backbone network pre-trained on ImageNet.
We trained one lite backbone network DS/64-12-16-1 on ImageNet, which obtains 66.8\% top-1 accuracy and 87.8\% top-5 accuracy on the validation-set (slightly worse than VGG-16).
After fine-tuning the whole detection framework on ``07+12" \texttt{trainval} set, we achieve 70.3\% mAP on the VOC 2007 test-set.
The corresponding training-from-scratch solution achieves 70.7\% accuracy, which is even slightly better.
Future work will investigate this point more thoroughly.

\vspace{-0.10in}
\subsubsection{Runtime Analysis}\label{running_time}
The inference speed is shown in the 6\emph{th} column of Table~\ref{VOC2007}.
With 300$\times$300 input, our full DSOD can process an image in 48.6ms (20.6 \textit{fps}) on a single Titan X GPU with the plain prediction structure,
and 57.5ms (17.4 \textit{fps}) with the dense prediction structure.
As a comparison, R-FCN runs at 90ms (11 \textit{fps}) for ResNet-50 and 110ms (9 \textit{fps}) for ResNet-101.
The SSD300$^*$ runs at 82.6ms (12.1 \textit{fps}) for ResNet-101 and 21.7ms (46 \textit{fps}) for VGGNet.
In addition, our model uses about only 1/2 parameters to SSD300 with VGGNet, 1/4 to SSD300 with ResNet-101, 1/4 to R-FCN with ResNet-101 and 1/10 to Faster R-CNN with VGGNet. {A lite-version of DSOD (10.4M parameters, w/o any speed optimization) can run 25.8 \textit{fps} with only 1\% mAP drops.}

\subsection{Results on PASCAL VOC2007}
Models are trained based on the union of VOC 2007 \texttt{trainval} and VOC 2012 \texttt{trainval} (``07+12'') following~\cite{liu2016ssd}. We use a batch size of 128.
Note that this batch-size is beyond the capacity of GPU memories (even for an 8 GPU server,  each with 12GB memory).
We use a trick to overcome the GPU memory constraints by accumulating gradients over two training iterations, which has been implemented on Caffe platform~\cite{jia2014caffe}.
The initial learning rate is set to 0.1, and then divided by 10 after every 20k iterations.
The training finished when reaching 100k iterations. Following~\cite{liu2016ssd}, we use a weight decay of 0.0005 and a momentum of 0.9.
All conv-layers are initialized with the ``xavier'' method~\cite{glorot2010understanding}.

Table~\ref{VOC2007} shows our results on VOC2007 \texttt{test set}. SSD300$^*$ is the updated SSD results which use the new data augmentation technique. Our DSOD300 with plain connection achieves 77.3\%, which is slightly better than SSD300$^*$ (77.2\%).
DSOD300 with dense prediction structure improves the result to 77.7\%. After adding COCO as training data, the performance is further improved to 81.7\%.
\begin{figure*}[]
	\centering
	\includegraphics[width=0.82\textwidth]{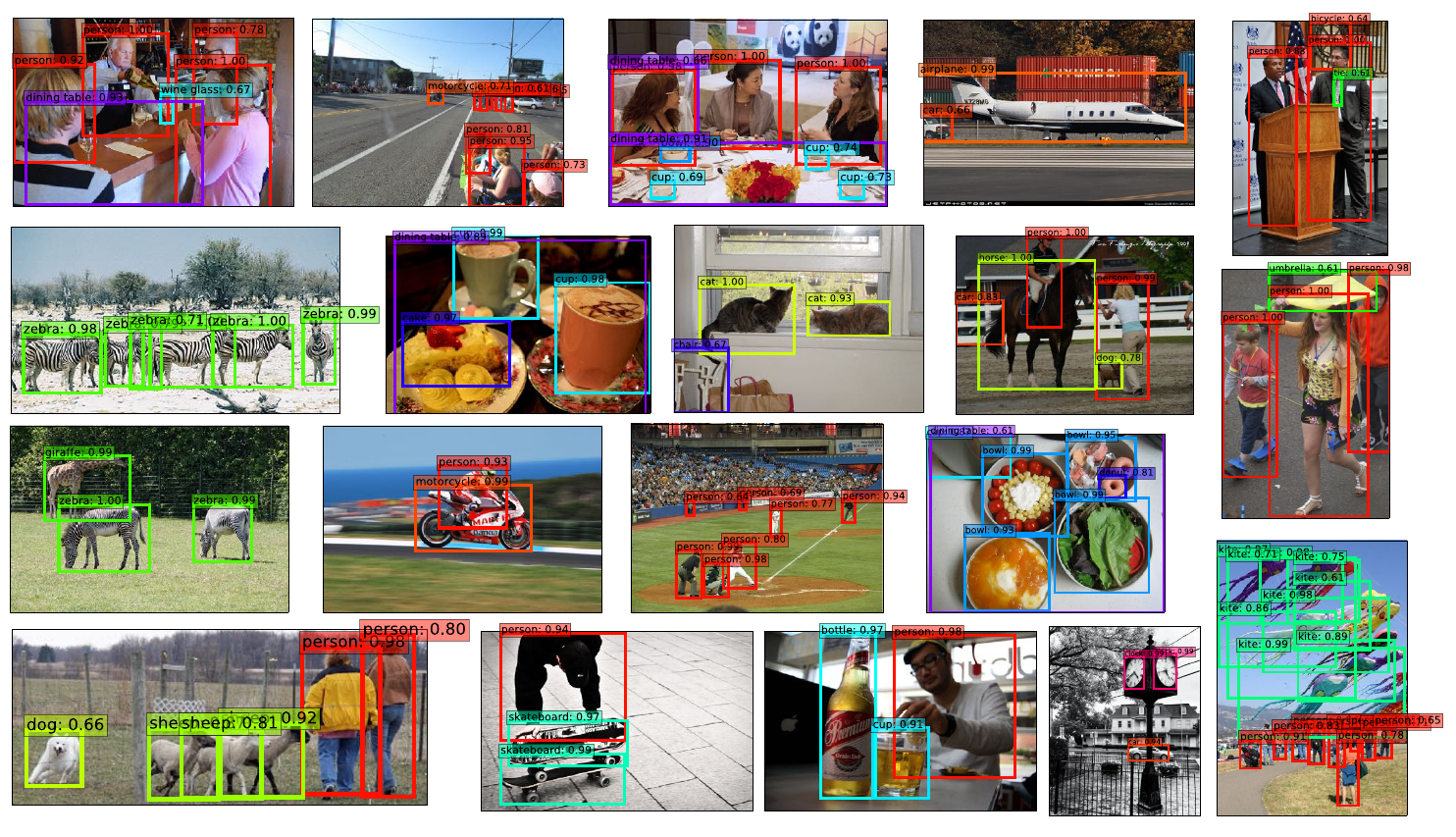}
	\vspace{-0.20in}
	\caption{Examples of object detection results on the MS COCO {\em {test-dev}} set using DSOD300. The training data is COCO {\em {trainval}} without the ImageNet pre-trained models (29.3\% mAP@[0.5:0.95] on the {\em {test-dev}} set). Each output box is associated with a category label and a softmax score in [0, 1]. A score threshold of 0.6 is used for displaying. For each image, one color corresponds to one object category in that image. The running time per image is 57.5ms on one Titan X GPU or 590ms on Intel (R) Core (TM) i7-5960X CPU @ 3.00GHz.}
	\label{examples}
	\vspace{-0.2in}
\end{figure*}

\subsection{Results on PASCAL VOC2012}
For the VOC 2012 dataset, we use VOC 2012 {\texttt {trainval}} and VOC 2007 {\texttt {trainval}} + {\texttt{test}} for training, and test on VOC 2012 {\texttt{test}} set. The initial learning rate is set to 0.1 for the first 30k iterations, then divided by 10 after every 20k iterations.
The total training iterations are 110k. Other settings are the same as those used in our VOC 2007 experiments.
Our results of DSOD300 are shown in Table~\ref{voc12}. DSOD300 achieves 76.3\% mAP, which is consistently better than SSD300$^*$ (75.8\%).

\subsection{Results on MS COCO}
Finally we evaluate our DSOD on the MS COCO dataset~\cite{lin2014microsoft}. MS COCO contains 80k images for training, 40k for validation and 20k for testing ({\texttt {test-dev}} set).
Following~\cite{ren2015faster,li2016r}, we use the {\texttt {trainval}} set (train set + validation set) for training.
The batch size is also set as 128.
The initial learning rate is set to 0.1 for the first 80k iterations, then divided by 10 after every 60k iterations. The total number of training iterations is 320k.

Results are summarized in Table~\ref{COCO}. Our DSOD300 achieves 29.3\%/47.3\% on the {\texttt {test-dev}} set, which outperforms the baseline SSD300$^*$ with a large margin. Our result is comparable to the single-scale R-FCN, and is close to the R-FCN{\footnotesize{multi-sc}} which uses ResNet-101 as the pre-trained model.
Interestingly, we observe that our result with 0.5 IoU is lower than R-FCN, but our [0.5:0.95] result is better or comparable. This indicates that our predicted locations are more accurate than R-FCN under the larger overlap settings. It is reasonable that our small object detection precision is slightly lower than R-FCN since our input image size (300$\times$300) is much smaller than R-FCN's ($\sim$ 600$\times$1000). Even with this disadvantage, our large object detection precision is still much better than R-FCN. This further demonstrates the effectiveness of our approach. Figure~\ref{examples} shows some qualitative detection examples on COCO with our DSOD300 model.

\section{Discussion}\label{discussion}
\paragraph{Better Model Structure {\em{vs.}} More Training Data.}
An emerging idea in the computer vision community is that object detection or other vision tasks might be solved with deeper and larger neural networks backed with massive training data like ImageNet~\cite{deng2009imagenet}.
Thus more and more large-scale datasets have been collected and released recently, such as the Open Images dataset~\cite{openimages},
which is 7.5x larger in the number of images and 6x larger of categories than that of ImageNet.
We definitely agree that, under modest assumptions that given boundless training data and unlimited computational power, deep neural networks should perform extremely well. However, our proposed approach and experimental results imply an alternative view to handle this problem: a better model structure might enable similar or better performance compared with complex models trained from large data.
Particularly, our DSOD is only trained with 16,551 images on VOC 2007, but achieves competitive or even better performance than those models trained with 1.2 million + 16,551 images.

In this premise, it is worthwhile rehashing the intuition that as datasets grow larger, training deep neural networks becomes more and more expensive. Thus a simple yet efficient approach becomes increasingly important. Despite its conceptual simplicity, our approach shows great potential under this setting.

\vspace{-2 ex}
\paragraph{Why Training from Scratch?}
There have been many successful cases where model fine-tuning works greatly.
One may ask why should we train object detectors from scratch.
We argue that, as aforementioned briefly, training from scratch is of critical importance at least for two cases.
First, there may be big domain differences from pre-trained model domain to the target one.
For instance, most pre-trained models are trained on large scale RGB image dataset, ImageNet.
It is very difficult to transfer ImageNet model to the domains of depth images, multi-spectrum images, medical images, etc.
Some advanced domain adaptation techniques have been proposed.
But what an amazing thing if we have a technique which can train object detector from scratch.
Second, model fine-tuning restricts the structure design space for object detection networks. This is very critical for the deployment of deep neural networks models to resource-limited Internet-of-Things (IoT) scenario.

\vspace{-2 ex}
\paragraph{Model Compactness {\em{vs.}} Performance.}
It has often been reported that there is a trade-off between model compactness (in terms of the number of parameters) and performance.
Most CNN-based detection solutions require a huge memory space to store the massive parameters.
Therefore the models are usually unsuitable for low-end devices like mobile-phones and embedded electronics.
Thanks to the parameter-efficient dense block, our model is much smaller than most competitive methods.
For instance, our smallest dense model (DS/64-64-16-1, with dense prediction layers) achieves 73.6\% mAP with only 5.9M parameters, which shows great potential for applications on low-end devices.

\section{Conclusion}\label{conclusion}
We have presented Deeply Supervised Object Detector (DSOD), a simple yet efficient framework for training object detector from scratch.
Without using pre-trained models on ImageNet, DSOD demonstrates competitive accuracy to state-of-the-art detectors such as SSD, Faster R-CNN and R-FCN on the popular PASCAL VOC 2007, 2012 and MS COCO datasets,
with only 1/2, 1/4 and 1/10 parameters compared to SSD,  R-FCN and Faster R-CNN, respectively.
DSOD has great potential on domain different scenario
like depth, medical, multi-spectral images, etc.
Our future work will consider these domains,
as well as learning ultra efficient DSOD models to support resource-bounded devices.

\begin{figure*}[]
	\centering
	\includegraphics[width=0.9\textwidth]{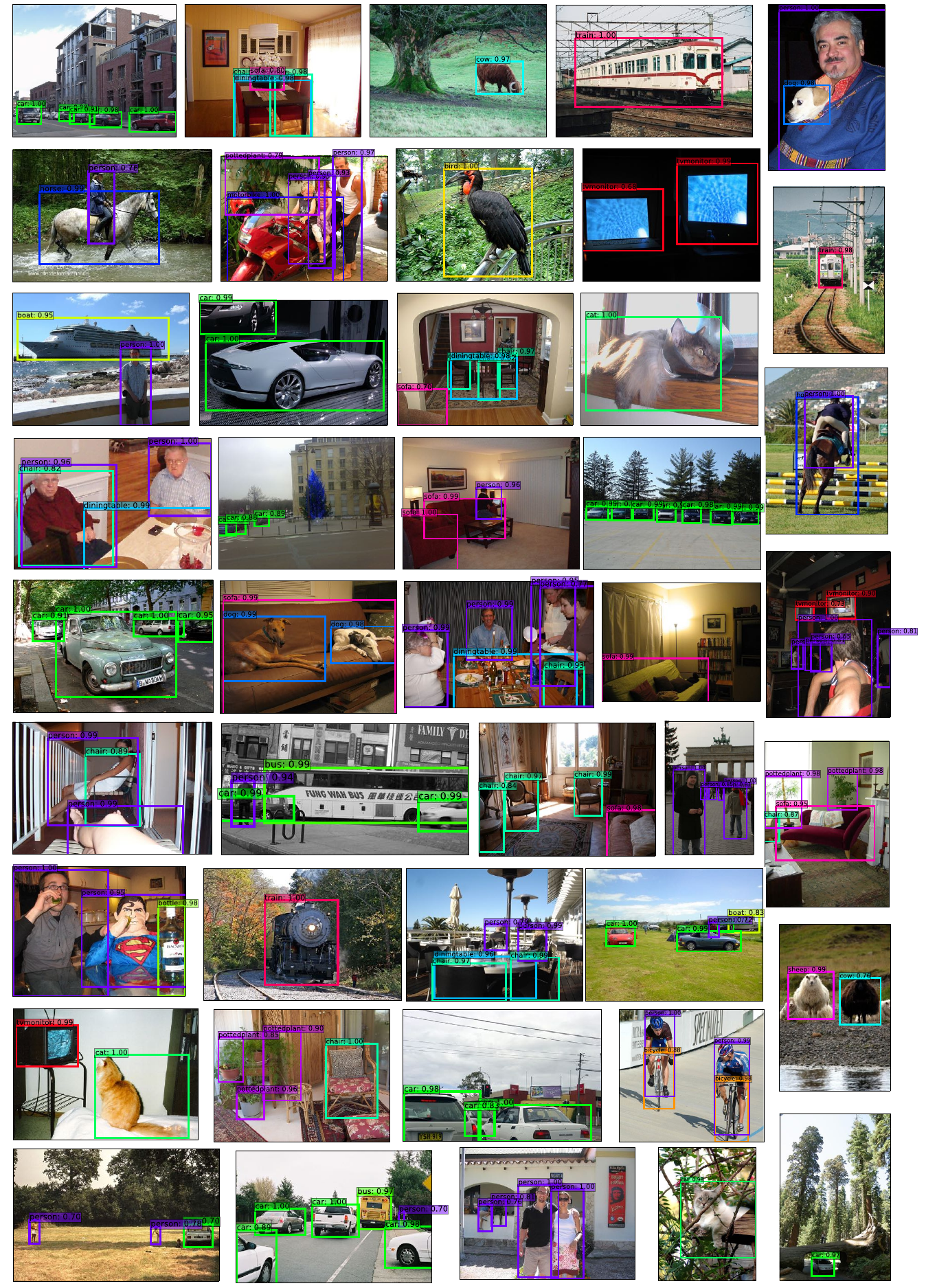}
	\vspace{-0.12in}
	\caption{More examples of object detection results on the PASCAL VOC 2007 {\em {test}} set using DSOD300. The training data is VOC 2007 {\em trainval}, VOC 2012 {\em trainval} and MS COCO {\em {trainval}} (81.7\% mAP@0.5 on the {\em {test}} set). Each output box is associated with a category label and a softmax score in [0, 1]. A score threshold of 0.6 is used for displaying. For each image, one color corresponds to one object category in that image.}
	\label{voc07}
	\vspace{-0.12in}
\end{figure*}

\begin{figure*}[]
	\centering
	\includegraphics[width=0.9\textwidth]{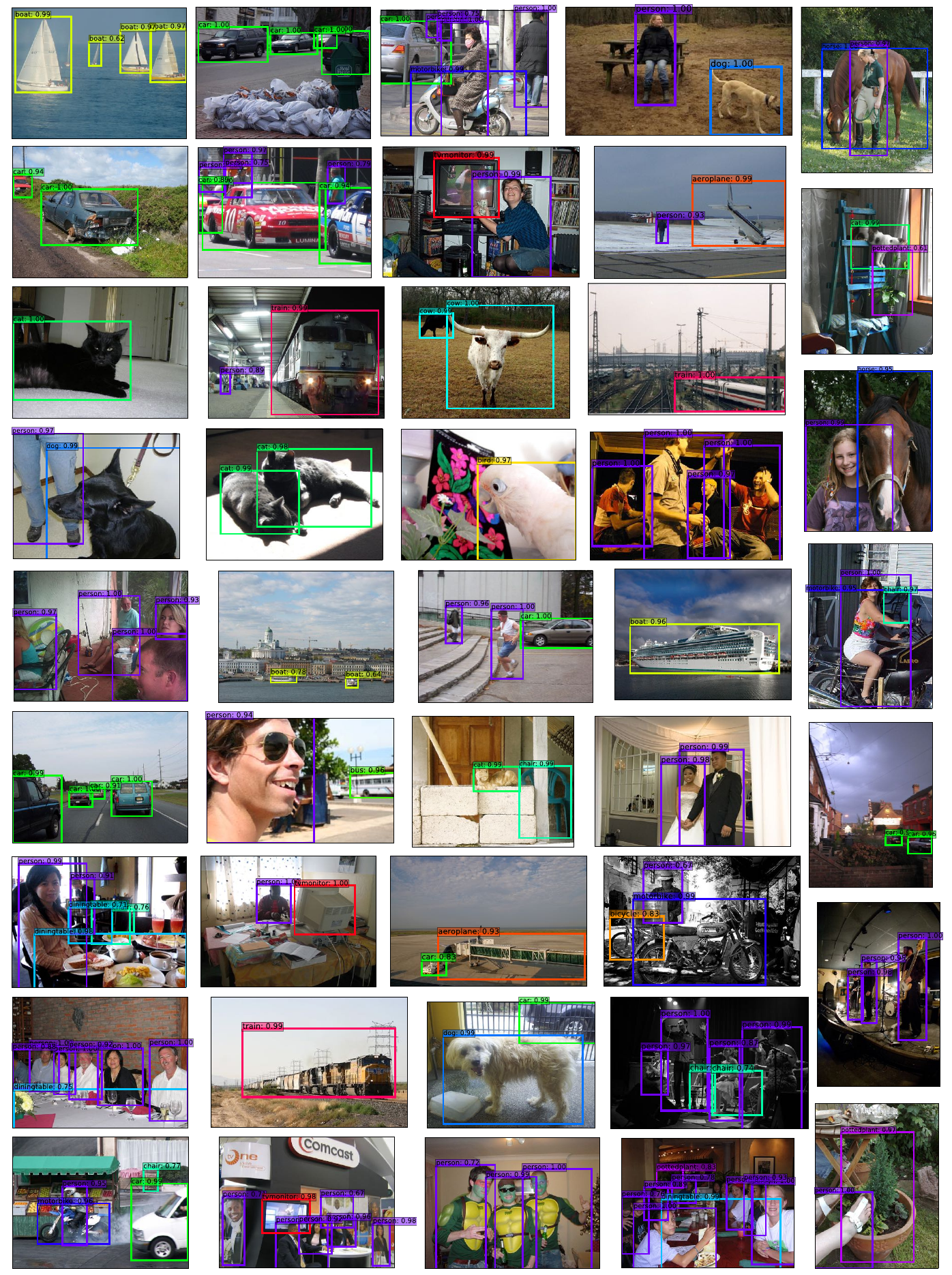}
	\vspace{-0.12in}
	\caption{More examples of object detection results on the PASCAL VOC 2012 {\em {test}} set using DSOD300. The training data is VOC 2007 {\em trainval}, VOC 2007 {\em test}, VOC 2012 {\em trainval} and MS COCO {\em {trainval}} (79.3\% mAP@0.5 on the {\em {test}} set). Each output box is associated with a category label and a softmax score in [0, 1]. A score threshold of 0.6 is used for displaying. For each image, one color corresponds to one object category in that image.}
	\label{voc12}
	\vspace{-0.12in}
\end{figure*}

\section*{Acknowledgements}
Yu-Gang Jiang and Xiangyang Xue are supported in part by a NSFC project (\#61622204), a project from STCSM (\#16JC1420401), and an European FP7 project (PIRSESGA-2013-612652).

   {\small 
	\bibliographystyle{ieee}
	\bibliography{egbib2}
}
\maketitle

\end{document}